\documentclass[hf]{ceurart}

\usepackage{listings}
\usepackage{footnote}
\usepackage{amsmath,amssymb,amsfonts}
\usepackage{algorithm}
\usepackage{algorithmic}
\usepackage{hyperref}
\usepackage{graphicx}
\usepackage{subfig}
\usepackage{textcomp}
\usepackage{xcolor}
\usepackage{subcaption}
\usepackage{tikz}
\usepackage{etoolbox}
\usepackage{caption}
\usepackage{tabularx}
\usepackage{multirow}

\usepackage{multicol}

\newsavebox{\boxA}
\newsavebox{\boxB}
\newsavebox{\boxC}

\begin{document}
\title{Can LLMs Translate Human Instructions into  a Reinforcement Learning Agent's Internal Emergent Symbolic Representation?}

\author[1]{Ziqi MA}[%
orcid=0009-0006-6775-4263,
email=ziqi_ma0605@163.com,
url=https://ziqilovesunshine.github.io,
]

\author[1]{Sao Mai NGUYEN}[%
orcid=0000-0003-0929-0019,
email=nguyensmai@gmail.com,
url=http://nguyensmai.free.fr,
]

\author[1]{Philippe XU}[%
orcid=0000-0001-7397-4808,
email=philippe.xu@ensta.fr,
url=https://perso.ensta-paris.fr/~philippe.xu,
]

\address[1]{U2IS, ENSTA, IP-Paris, 828 Bd des Maréchaux, 91120 Palaiseau, France}

\copyrightyear{2022}
\copyrightclause{Copyright for this paper by its authors.
  Use permitted under Creative Commons License Attribution 4.0
  International (CC BY 4.0).}

\conference{1st Workshop on Interactive Task Learning in Human-Robot co-construction (ITL4HRI) | IEEE RO-MAN 2025: International Conference on Robot and Human Interactive Communication 2025, August 25–29, 2025, Eindhoven, Netherlands}

\tnotemark[1]
\tnotetext[1]{The work is partially supported by Hi! Paris.}


\begin{abstract}
Emergent symbolic representations are critical for enabling developmental learning agents to plan and generalize across tasks. In this work, we investigate whether large language models (LLMs) can translate human natural language instructions into the internal symbolic representations that emerge during hierarchical reinforcement learning. We apply a structured evaluation framework to measure the translation performance of commonly seen LLMs --
GPT, Claude, Deepseek and Grok -- across different internal symbolic partitions generated by a hierarchical reinforcement learning algorithm in the Ant Maze and Ant Fall environments. Our findings reveal that although LLMs demonstrate some ability to translate natural language into a symbolic representation of the environment dynamics, their performance is highly sensitive to partition granularity and task complexity. The results expose limitations in current LLMs capacity for representation alignment, highlighting the need for further research on robust alignment between language and internal agent representations.
\end{abstract}
\begin{keywords}
  Emergent Agent \sep
  Symbolic Representation\sep
  STAR algorithm \sep
  LLMs \sep
  alignment capacity \sep
\end{keywords}

\maketitle

\section{Introduction}

Emergent symbolic representation refers to the spontaneous development of internal symbols or abstract concepts within a learning agent, without explicit pre-programming of those symbols but during learning processes that involve interaction with the environment, such as reinforcement learning (RL) or human-in-the-loop learning paradigms~\cite{Retzlaff2024JAIR}. While most works have focused on symbol emergence for primitive actions, few works have investigated for long-horizon actions, whether the emergent symbols~\cite{Zadem20232IICDLI,Ahmetoglu2022JAIR} are aligned with human representations. 
``Symbol emergence is a critical concept for understanding and creating cognitive developmental systems that are capable of behaving adaptively in the real world and that can communicate with people'', as highlighted in~\cite{Taniguchi2018ITCDS}. The authors in~\cite{akakzia2020grounding} have demonstrated that language instructions can be integrated with emergent symbolic representations of an RL agent in a end-to-end manner. In this work, we investigate the importance of emergent symbols in developmental learning agents to communicate with people. How do these two sets of symbols that have a prior different representations align? What can be used to bridge the gap between internal symbols of developmental learners and natural language symbols used by human instructors? 

Large Language Models (LLMs) have emerged as a powerful tool in natural language processing, demonstrating remarkable capabilities in understanding and generating human language~\cite{naveed2023comprehensive} and to encode common-sense knowledge. They have become advanced artificial intelligence systems capable of understanding~\cite{y2022large}, generating, and generalizing text across various tasks~\cite{radford2019language, brown2020language}. 
Recently, LLMs have shown reasoning and inferring capacity~\cite{qian2023chatdev, yao2023tree}, which improve the agent decision-making ability. 
While the research community has investigated the ability of LLMs to understand human language, it has overlooked investigating the ability of LLMs to understand agent "language" which is the internal state representation of an agent during the learning process, and the ability of LLMs in aligning the internal representation of humans and the internal representation of an artificial learning agent of the same environment. In our paper, we focus on the following question: for an emergent symbolic representation in developmental learning agents, can LLMs translate human instructions from natural language into the internal representation of the agent? In order to explore the question, we analyze the performance of LLMs to align human instructions in natural language with the dynamic symbolic representation developed by a reinforcement learning agent. Our work is based on the hierarchical reinforcement learning algorithm Spatio-Temporal Abstraction Via Reachability (STAR)\cite{Zadem2024TICLR}, which was shown to learn a symbolic representation of the state space \cite{Zadem20232IICDLI} grounded on its environments. 
The main contributions of our paper are:
\begin{itemize}
    \item We show that LLMs can be a tool to translate human natural language instructions into the internal representation of STAR during its learning process.
    \item We show that the capacity of LLMs to align two different representations is limited due to the fact that they do not fully leverage the internal representation learned by the developmental agent.
\end{itemize}

We state the preliminary of our research in section \ref{sec:preliminary}, propose the method in section \ref{sec:methodology}, and describe the implementation details in section \ref{sec:implementation}. Then, we design the experiments and analyze the results in section \ref{sec:experiment}. Finally, we discuss the result and give our conclusion in section \ref{sec:dicussion} and section \ref{sec:conclusion}.

\begin{figure}
  \centering
  \includegraphics[width=\linewidth]{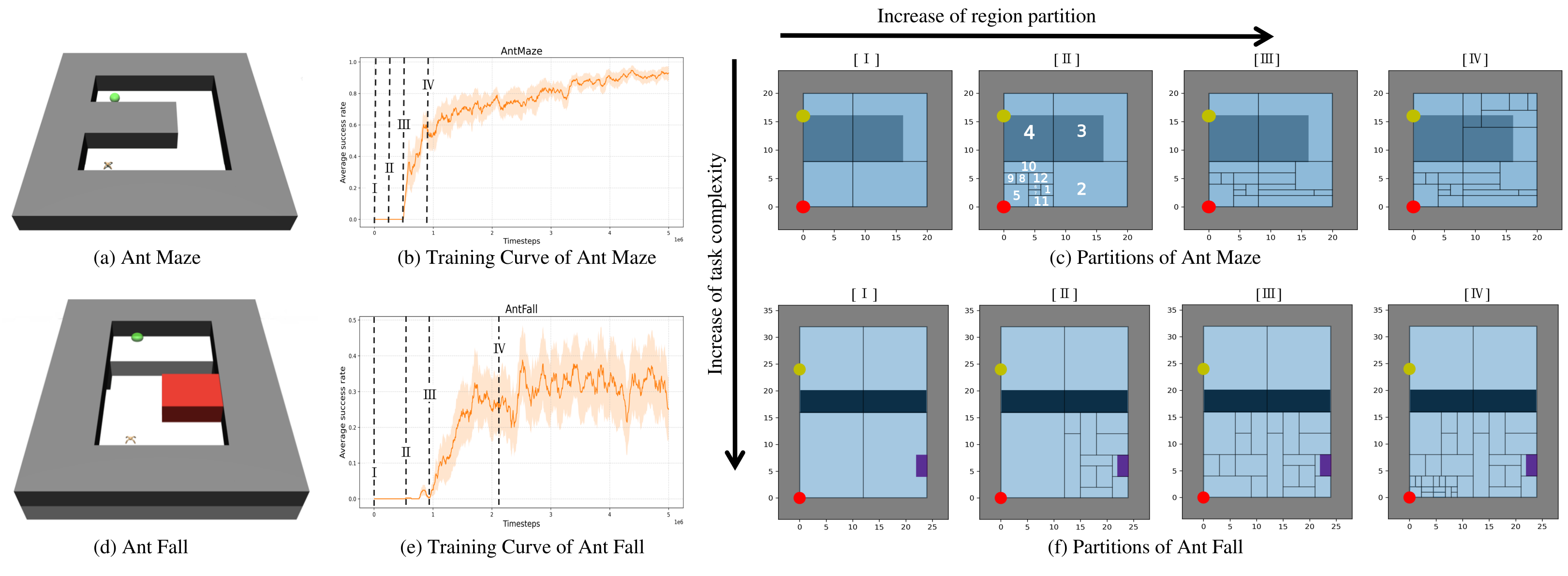}
  \caption{(a)(d) Environments, (b)(e) Average success rate of STAR  (from \cite{Zadem2024}), (c)(f) Partition into regions of STAR, in respectively Ant Maze and Ant Fall. The regions in (c)(f)  are the internal representation emerging during the training at timestamps noted in (b) and (e). The red point represents the initial position of the robot while the yellow point represents the goal position. Our LLM-based system translates  instructions to guide the robot (such as "go east to the end, turn north until past the wall and go west until the end"), into a sequence of traversed regions (for Partition II of AntMaze, the output is 5 → 11 → 2 → 3 → 4).}
\label{fig:partitions_antmaze_antfall}
\end{figure}

\section{Preliminary}
\label{sec:preliminary}
We base our work on running STAR at Ant Maze\cite{duan2016benchmarking} and Ant Fall\cite{nachum2018data} tasks. STAR is a 3-layers Hierarchical Reinforcement Learning algorithm aiming to solve long-horizon tasks: the high-level agent selects regions in an abstract goal space, the middle-level agent selects concrete subgoals that aid in achieving abstract goals, and the low-level agent learns to take action in the environment. STAR automatically discovers discrete symbolic representations of environment by grouping states with similar reachability properties into symbolic goal regions and allowing incremental refinement. The abstract symbols evolve over time as new experiences improve the learned representation. When running STAR on one task, a set of internal symbolic representation data $S=\{s_1, s_2, ..., s_N\}$ emerges. Based on the mechanism of STAR, $s_i$ represents a set of emergent internal symbols that align with environment dynamics, and the $s_{i+1}$ is built on the top of $s_{i}$.

We collect the symbolic partitions that emerge during the execution of the STAR algorithm on the Ant Maze and Ant Fall environments, as illustrated in Fig.~\ref{fig:partitions_antmaze_antfall}(a) and (d). The Ant Maze environment features a quadrupedal robotic agent navigating through a “$\supset$”-shaped corridor toward a fixed goal location, and the Ant Fall task simulates an environment where the robot is initially placed on a raised platform separated by a pit from its target. The agent must learn that it needs to navigate to the movable block, push it into the pit to form a bridge, and then cross the gap to reach the target. During the training process, STAR incrementally refines its symbolic representation. To analyze LLM performance across different levels of abstraction, we select four representative partitions from different developmental learning stages, as shown in Fig.\ref{fig:partitions_antmaze_antfall}(c) and (f). Their corresponding positions are also marked along the training curves in Fig.\ref{fig:partitions_antmaze_antfall}(b) and (e). Partition I corresponds to the initialization partition with a minimal number of symbols; Partition II captures a timestep before any significant learning progress; Partition III aligns with the onset of performance improvement;  Partition IV represents final stage of learning. Let us note that Ant Fall is more complex than Ant Maze as it includes an additional pushing task.


\section{Methodology}
\label{sec:methodology}

\begin{algorithm}[ht!]
\caption{Translation by LLM of Human Instructions into Emergent Symbolic Representations}
\label{alg:methodology}
\begin{algorithmic}
\REQUIRE  $S = \{s_1, s_2, \ldots, s_N\}$: symbolic representations from STAR,\\
$I = \{I_{1}, I_{2}, \ldots, I_{N}\}$: set of human instructions sets,\\
$G = \{G_{1}, G_{2}, \ldots, G_{N}\}$: set of ground truth output sets
\ENSURE $M = \{M_{1,1}, M_{1,2}, \ldots, M_{N,J}\}$: evaluation scores
\FOR{$i = 1$ to $N$}
    \FOR{$j = 1$ to $J$}
        \STATE $p_{i,j} \gets f_{\text{prompt}}(s_i, I_{i, j})$
        \FOR{$k = 1$ to $K$}
            \STATE $o_{i,j,k}^{\text{LLM}} \gets \text{LLM}(p_{i,j}, r_k)$
            \STATE $m_{i,j,k} \gets \max_{q=1\ldots Q} M\left(o_{i,j,k}^{\text{LLM}}, G_{i,j,q}\right)$
        \ENDFOR
        \STATE $M_{i,j} \gets \frac{1}{K} \sum_{k=1}^{K} m_{i,j,k}$
    \ENDFOR
\ENDFOR
\end{algorithmic}
\end{algorithm}

Let $S = \{s_1, s_2, \ldots, s_N\}$ denote the set of internal symbolic representations identified during the developmental learning process of STAR. We design a collection of instructions $I = \{I_1, I_2, \ldots , I_N\}$ from $J$ humans, where each element is a set of instructions $I_i = \{I_{i,1}, I_{i,2}, \ldots, I_{i,J}\}$. Each instruction $I_{i,j}$ corresponds to a unique natural language command describing the goal or behavior related to $s_i$ by the human $j$. We define a prompt construction function $f_{\text{prompt}}(s_i, I_{i,j})$ that takes a symbolic representation $s_i$ and an associated instruction $I_{i,j}$ to generate a textual prompt $p_{i,j}$ for the LLM:
\begin{align}
p_{i,j} &= f_{\text{prompt}}(s_i, I_{i, j})
\end{align}

Due to the stochastic nature of LLMs, a given prompt $p_{i,j}$ may result in different outputs across multiple queries. We introduce a random state $r_k$ (\textit{e.g.}, random seed) and define the LLM-generated output at query time $k$ as:
\begin{align}
o_{i, j, k}^{LLM} &= \text{LLM}(p_{i, j}, r_k)
\end{align}

To establish a reference for evaluation, domain experts provide human-annotated ground truth outputs $G_{i,j}$ for each symbolic-instruction pair $(s_i, I_{i,j})$, so for each $G_i \in G$, $ G_i = \{G_{i1}, G_{i2}, ... , G_{iJ}\}$.
It is possible that multiple reference outputs are compatible with the pair $(s_i, I_{i,j})$, in this case we consider a set of references  $G_{ij} = \left\{G_{i,j,1},\ldots,G_{i,j,Q}\right\}$ for each $G_{i,j} \in G_i$. The set of ground truth output sets is the collection of all these references:  $G = \left\{G_1, G_2, \ldots G_N\right\}$.
We then define an evaluation metric $M\left(o^{\text{LLM}}, G\right)$ that measures the similarity between the LLM-generated output and the corresponding human-provided reference. Since there may exist several possible ground truth responses associated with one internal symbolic set and one instruction, we compute the performance for each pair by taking the maximum similarity across all human-provided references $G_{i,j,q}$. The similarity is high if at least one of the references provides a high similarity. The average score over $K$ runs is defined as:
\begin{align}
M_{i,j} &= \frac{1}{K} \sum_{k=1}^{K} \max_{q=1\ldots Q} M\left(o_{i, j, k }^{\text{LLM}}, G_{i, j, q}\right)
\label{eq:m}
\end{align}

This formulation allows us to robustly evaluate the performance of LLMs to translate symbolic representations into interpretable language aligned with human expectations. The overall algorithmic procedure is described in Algorithm~\ref{alg:methodology}.

\section{Implementation}
\label{sec:implementation}

\subsection{Designs of Prompt}
In order to reduce the ambiguity and optimize the information processing of LLM, the prompt that is given to the LLM needs to be carefully designed. A well-structured prompt guides the model toward producing outputs that align with the intended task or reasoning process. Considering the characteristic of the algorithm STAR, we design a prompt that includes four terms:
\begin{itemize}
    \item Graph-Based Representation: We explicitly map the internal representation into a graph which encourages the agent to navigate using abstracted region connections.
    \item Explicit State and Goal Representation: We represent the current state and goal symbolically on the graph.
    \item Instruction-to-Graph Mapping: We translate human language into a structured symbolic reasoning process, ensuring alignment between linguistic commands and executable movements, i.e. actions to move to a neighbour region.
    \item Final output form: We regularize the final output form to facilitate the evaluation process.
\end{itemize}

Examples of prompts in Ant Maze and Ant Fall are given in Appendix \ref{appendix-prompt}

\subsection{Selection of Evaluation Metrics}
Since our task consists in translating a natural language sentence into a sequence of symbols emerged in the robot representing a region traversal sequence, order and precision are critical. Its evaluation requires a metric that penalizes incorrect orderings while tolerating minor structural variations. Google-BLEU (G-BLEU) \cite{wu2016googles} is an improved version of BLEU \cite{Papineni02bleu, lin-och-2004-orange},which is the quotient of the matching words under the total count of words in hypothesis sentence (traduction). Regarding the denominator BLEU is a precision oriented metric. G-BLEU 
applies a smoother brevity penalty and better handles sparse n-gram statistics. 
By rewarding partial correctness and preserving order sensitivity, G-BLEU is well-suited for assessing the alignment between LLM-generated outputs and symbolic plans. The evaluation metric $M$ in~\eqref{eq:m} used is G-BLEU. As this study does not aim for autonomous navigation, already assessed by the algorithm STAR, we do not report success rate of the task.


\section{Experimental Evaluation}
\label{sec:experiment}

\subsection{Translation Ability of LLMs Across Partition Granularity}
\label{subsec:tran LLM}

\begin{table}[ht]
\caption{Mean G-BLEU scores of LLMs over 4 runs for each partition in the Ant Maze and Ant Fall environments.}
\label{tab-llm_performance}
\centering
\begin{tabular}{|l|c|c|c|c||c|c|c|c|}
\hline
\multirow{2}{4em}{\textbf{LLMs}} & \multicolumn{4}{c||}{\textbf{Ant Maze}} & \multicolumn{4}{c|}{\textbf{Ant Fall}} \\
\cline{2-9}
            & P-I & P-II & P-III & P-IV & P-I & P-II & P-III & P-IV \\
\hline
\hline
GPT o3-m  & 1    & 1     & 1      & 0.87  & 1    & 0.80  & 0.73   & 0.86\\
Claude  & 1    & 1     & 0.73 & 0.34  & 1    & 0.50  & 0.62   & 0.77 \\
Deepseek    & 1    & 0.9   & 0.53   & 0.65  & 1    & 0.66  & 0.74   & 1 \\
GROK        & 1    & 1     & 1      & 0.89  & 1    & 0.66     & 0.74   & 1 \\
\hline
\end{tabular}
\end{table}

In the first experiment, we evaluate the consistency of LLMs in interpreting fixed natural language instructions across varying symbolic abstractions. For both the Ant Maze and Ant Fall environments, we keep the agent's start and goal positions fixed and apply the same instruction to all partition levels. The instruction for Ant Maze is: “Move right until you completely pass the wall on your left, move up until you have crossed the upper wall, turn left and proceed until you reach the goal.” For Ant Fall it is: “Move east until reaching the end, then go north to push the movable block into the pit, cross the pit using the block as a bridge, and finally head west to reach the goal.”

We report in Table~\ref{tab-llm_performance} the G-BLEU score for the translation of these instructions into a sequence of regions by four commonly used reasoning LLMs: GPT o3-mini, Claude 3.7, DeepSeek-r1, and GROK. 
While GPT o3-mini achieves the highest scores, all LLMs achieve scores above 0.5 across tasks, indicating a generally successful translation of human instructions into the agent’s internal symbolic representation. In the Ant Maze environment, all LLMs scores decrease as the number of regions increases. This observation suggests that as the symbolic space becomes more fine-grained, the increase in abstraction complexity challenges the LLMs’ ability to produce coherent and accurate language-to-symbol translation. Notably, the degradation of performance varies by model: GPT o3-mini and GROK demonstrate greater robustness than DeepSeek-r1 and Claude 3.7. In Ant Fall, however, the translation performance follows a non-monotonic trend—initially decreasing and then partially recovering. This pattern is likely due to the task’s increased complexity, which involves not only spatial navigation but also object manipulation. 

Given the consistent trends observed across LLMs, we select GPT o3-mini as the representative LLM for subsequent experiments. Furthermore, to better characterize the translation performance in Ant Fall, we segment the task into two phases: (1) before the block, where the agent must reach the object, in a navigation task similar to AntMaze; and (2) after the block, where the agent has to use the block as a tool to reach the goal. This separation can highlight the LLM translation behavior under 
the environment change induced by tool use.

{\captionsetup[subfloat]{captionskip=0pt}
\begin{figure*}[!h]
\centering
\sbox{\boxA}{\subfloat[Ant Maze]{\includegraphics[width=2.2in]{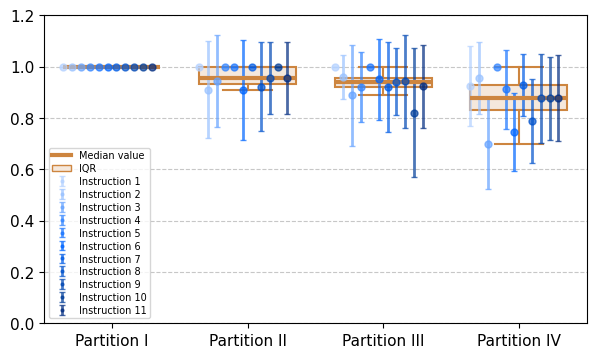}}}
\sbox{\boxB}{\subfloat[Ant Fall, before block]{\includegraphics[width=2.2in]{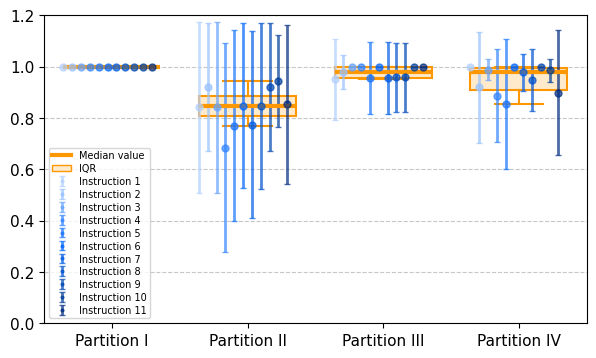}}}
\sbox{\boxC}{\subfloat[Ant Fall, after block]{\includegraphics[width=2.2in]{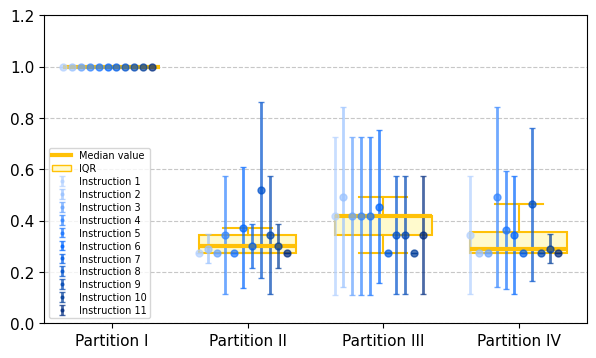}}}

\begin{tikzpicture}
  \node (C) at (10.36,0) {\usebox{\boxC}};
  \node (B) at (5.18,0) {\usebox{\boxB}};
  \node (A) at (0,-0.16) {\usebox{\boxA}};
\end{tikzpicture}
\caption{G-BLEU scores for partition-agnostic instructions tested in (a) Ant Maze, and (b) Ant Fall before block and (c) Ant Fall after block. For each internal representation, we plot in blue the average and standard deviation of 10 queries for each instruction, and boxplot in brown, orange and yellow the average and IQR over the 11 instructions.}
\label{well_design}
\end{figure*}
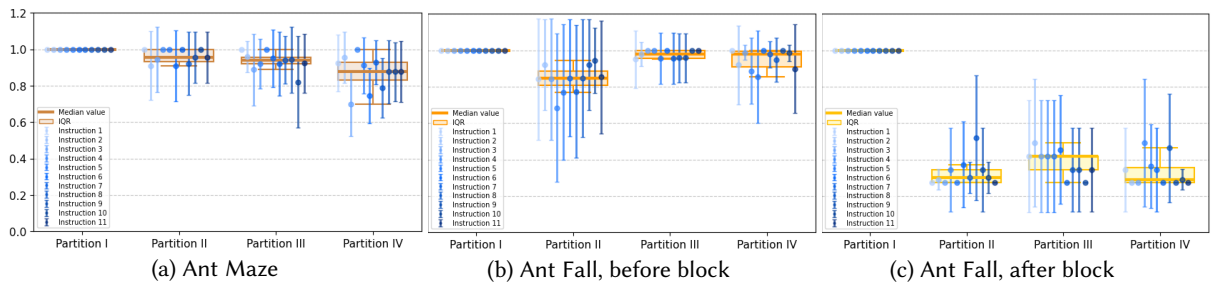
}

In order to test the robustness of the previous results with statistical tests, 
we design 11 different instructions for each environment (shown in Appendix \ref{11instructions}), and tested all 11 instructions on all four partitions. We constructed the prompt with the method introduced in Section~\ref{sec:methodology} and queried the LLM 10 times. Figure~\ref{well_design} illustrates the average G-BLEU scores 
across the symbolic partitions of Ant Maze, Ant Fall before the block and Ant Fall after the block. We observe a perfect translation performance in Partition I both in Ant Maze and Ant Fall. The trend of translation performance observed in Ant Maze is a consistent slightly drop from Partition I to Partition IV, while in Ant Fall before block, the translation performance drops from Partition I to Partition II, followed by a recovery or stabilization in Partitions III and IV. This trend is interpretable through the partition structures shown in Figure~\ref{fig:partitions_antmaze_antfall}. Partition I represents a very coarse abstraction with minimal region division—making it easier for the LLM to infer plausible region sequences regardless of the instruction quality. Then, when the partition becomes more granular in Ant Maze, the LLM is more prone to  mistakes. In Ant Fall before Block, however, in Partition II, the bottom-left is partitioned as a single big region, while the bottom-right is partitioned into small regions. This introduces a confusion : which small region of the right corresponds to the displacement to the right from the big bottom-left region? In contrast, in Partition III, the bottom-left part becomes finer, which lifts this ambiguity. In Partition IV, more regions emerge in the already partitioned bottom-left part, which does not introduce too much change to the performance. On the contrary, in Ant Fall after Block, we observe low scores from Partition II, III and IV. This part includes the agent's use of the movable block as a tool to build a route: after the agent moves the block, the environment changes, and the LLM fails to capture the change in the environment, thus fails to translate the instructions. These results indicate that the LLM has some ability to translate the human instruction into internal symbolic representation from coarse to fine, while it has some difficulties to manage tool use. 


\subsection{Variability in the Instructions}

{\captionsetup[subfloat]{captionskip=0pt}
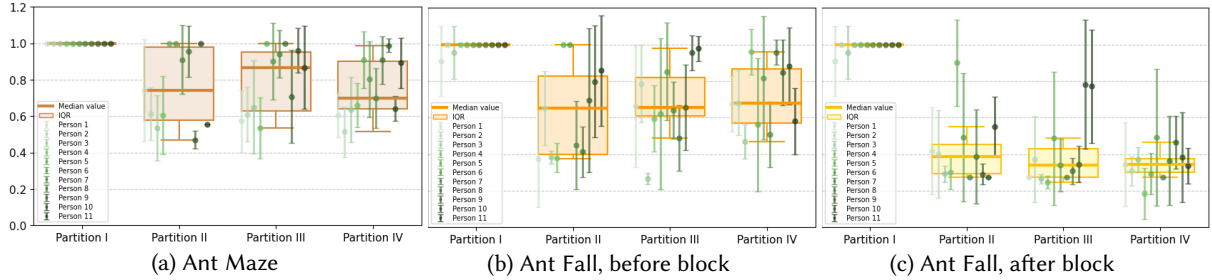
\begin{figure*}[!t]
\centering
\sbox{\boxA}{\subfloat[Ant Maze]{\includegraphics[width=2.2in]{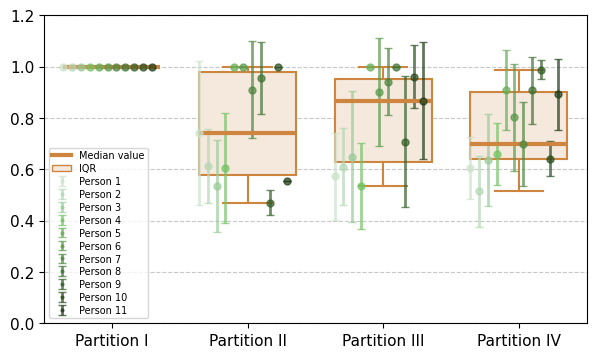}}}
\sbox{\boxB}{\subfloat[Ant Fall, before block]{\includegraphics[width=2.2in]{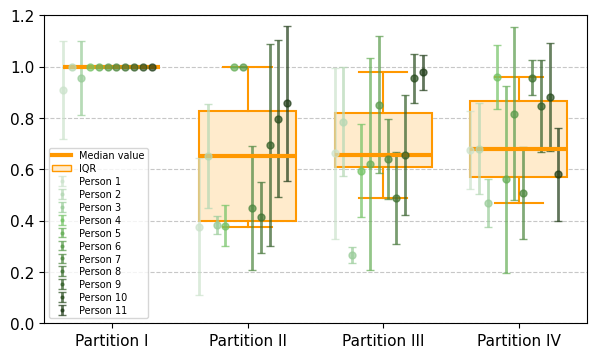}}}
\sbox{\boxC}{\subfloat[Ant Fall, after block]{\includegraphics[width=2.2in]{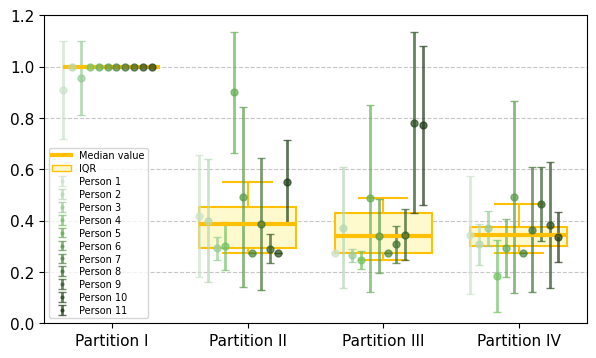}}}

\begin{tikzpicture}
  \node (C) at (10.36,0) {\usebox{\boxC}};
  \node (B) at (5.18,0) {\usebox{\boxB}};
  \node (A) at (0,-0.16) {\usebox{\boxA}};
\end{tikzpicture}
\vspace{-1.5em}
\caption{G-BLEU scores for partition-associated instructions  tested for each internal representation in (a) Ant Maze, (b) Ant Fall before block and (c) Ant Fall after block. For each tested internal representation, we plot in green the average and standard deviation of 10 queries for instruction from each person, and we show the median and the IQR over the 11 persons by boxplot.}
\label{translation_ability}
\vspace{-1.em}
\end{figure*}
}

To test the robustness of our results to the variability of instructions, we asked 11 participants most of whom are researchers in the lab, but not knowledgeable about hierarchical reinforcement learning or the algorithm STAR, to write their own verbal instructions in the Ant Maze and Ant Fall environments. 
We created a questionnaire in which the participants were shown the partition layouts in Fig.\ref{fig:partitions_antmaze_antfall}(c)(f) and were asked to express a general instruction: “go →, then go ↑, and finally go ← to reach the goal,” by using their own words, with instructions that correspond to the region partitions and to differentiate their descriptions for each partition. 
 The average G-BLEU scores between the LLM-generated symbolic output and the ground truth, over ten runs of the LLM, are presented in Fig.\ref{translation_ability}. The higher variance in Fig. \ref{translation_ability} than in Fig. \ref{well_design} is due to the instructions originating from 11 different participants outside of the project, instead of a single person. 

At Ant Maze Partition-I, G-BLEU scores are consistently close to 1.0, indicating high overlap between the LLM-generated region sequence and the ground truth sequence. This reflects accurate interpretation and planning under a simple abstraction. As partition granularity increases, G-BLEU scores decline, with growing variance across participants. The increased fragmentation of regions introduces more possible region transitions, leading to higher risk of deviation in LLM outputs. For Ant Fall, the performance of LLM for partition-associated instructions drops a little by comparing to Fig.\ref{well_design}, however the tendency does not change. This indicates the conclusion that we got from Section \ref{subsec:tran LLM} is valid.


\subsection{Assessing Representation Alignment via Instruction-Partition Mismatch}

We evaluate how well an LLM aligns natural language instructions in the internal symbolic spatial representations by testing instruction–environment mismatch scenarios in Ant Maze and Ant Fall. We analyze two complementary setups: (1) applying instructions gathered for the simplest partition to all partition levels, and (2) applying instructions gathered for the most complex partition to all levels. 

Figure \ref{simpliest-AntMaze} to \ref{simpliest-AntFallAfterB} plot the G-BLEU of the simplest partition instructions applied across all levels in Ant Maze task and Ant Fall. We find that instructions corresponding to the simplest partition yields nearly perfect language alignment on the simplest partition level. However, as the same coarse-partition instruction is applied to more complex partitions, performance degrades and variability increases. The high variation between participants underscores that the LLM translation of the simplest partition instructions in a very complex environment is highly inconsistent. We also notice that the decline is not strictly linear with partition complexity. In Ant Fall, once these instructions are applied to more complex partition levels, the performance drops markedly, more so than in Ant Maze. Interestingly, the trend is not strictly monotonic: at Ant Fall Partition-III the median rises slightly, recovering some performance even though the partition mismatch is larger. This non-monotonic behavior suggests that the relationship between partition granularity mismatch and G-BLEU score is complex. By comparing Fig.\ref{simpliest-AntMaze} to \ref{simpliest-AntFallAfterB} with Fig.\ref{translation_ability}, we observe that an eminent increase in all figures, which totally contradicts what we expect: The G-BLEU scores drop sharply in a properly aligned-but-mismatched scenario. This failure pattern illustrates that the LLM struggle to maintain symbolic translation ability from human natural language: instructions refer to certain regions can erroneously be repurposed by the model in a different context, because the model does not accurately “understand” region partitions as internal representations.

Figure \ref{complex-AntMaze} to \ref{complex-AntFallAfterB} show the results of the instructions associated with the most complex partition applied to all partitions. Intuitively, one would expect the performance to be the highest when these detailed instructions are used on their corresponding fine-grained partition. However, contrary to expectations, the results show an almost inverted pattern. On the associated complex partition, the G-BLEU scores are only moderate, this suggests that the complex partition instructions themselves may be poorly learned and the LLM lacks the capacity of alignment of two representations even in the ideal case. When the same complex instructions are applied on the simplest partition, the G-BLEU scores remain exceptionally high, which corresponds to the performance that we observe in Fig.\ref{well_design}. This is because the partition is too simple to make a mistake, so LLM defaults to describing the high-level route correctly. When we focus on the intermediate partitions and compare them to the associated-instruction performance in Fig.\ref{translation_ability}, the no-dropping scores of Partition-II and Partition-III further reflect that the LLM is not leveraging the symbolic region structure as intended.
It points to conclude that the translation ability of LLM relies on some surface-level patterns and is not firmly aligned on the understanding of internal symbolic representation.

Additionally, the inter-participant variability is significant in the experiments, we see some runs apparently stumbling on instructions while others fail. It suggests the LLM behavior is somewhat random with respect to the symbolic alignment. In particular,  participants 1, 2, 11 mentioned the partitions explicitly in their descriptions, for instance, the instruction of participant 11 is "Move right across three regions, then move up across five regions, move left at the end of the current region, then move down until the obstacle, and finally move left until the goal." However, the change of the G-BLEU score for the instructions applied on the different partitions is not correlated to whether they are aligned or not.

{\captionsetup[subfloat]{captionskip=0pt}
\begin{figure*}[!t]
\centering
\sbox{\boxA}{\subfloat[Simple, Ant Maze]
{\includegraphics[width=2.2in]{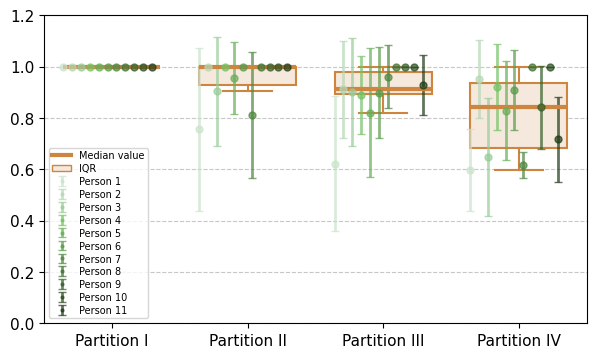}
\label{simpliest-AntMaze}}}
\sbox{\boxB}{\subfloat[Simple, Ant Fall, after block]{\includegraphics[width=2.2in]{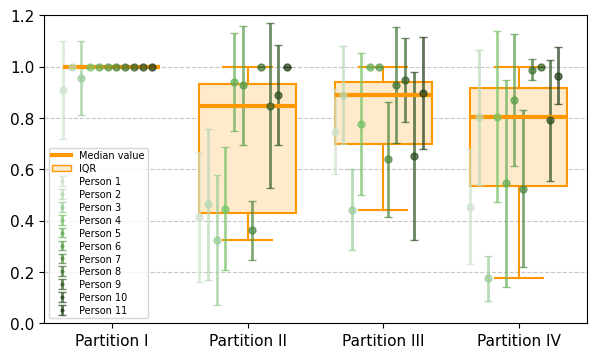}}}
\sbox{\boxC}{\subfloat[Simple, Ant Fall, after block]{\includegraphics[width=2.2in]{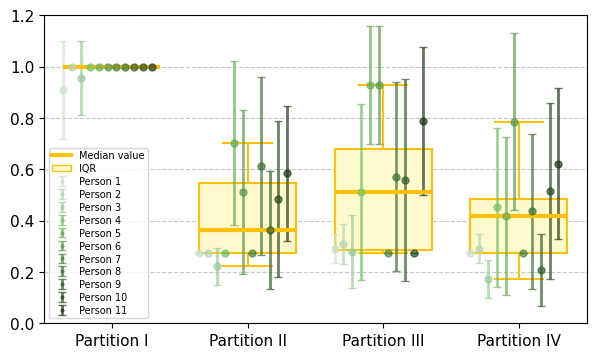}
\label{simpliest-AntFallAfterB}}}

\begin{tikzpicture}
  \node (C) at (10.36,0) {\usebox{\boxC}};
  \node (B) at (5.14,0) {\usebox{\boxB}};
  \node (A) at (0,-0.16) {\usebox{\boxA}};
\end{tikzpicture}
\vspace{-0.5em}
\caption{G-BLEU scores for Instructions specific to the simplest partition applying across all partitions in (a) Ant Maze task, (b) Ant Fall task, before block, (c) Ant Fall task, after block.}

\vspace{-1.em}
\end{figure*}
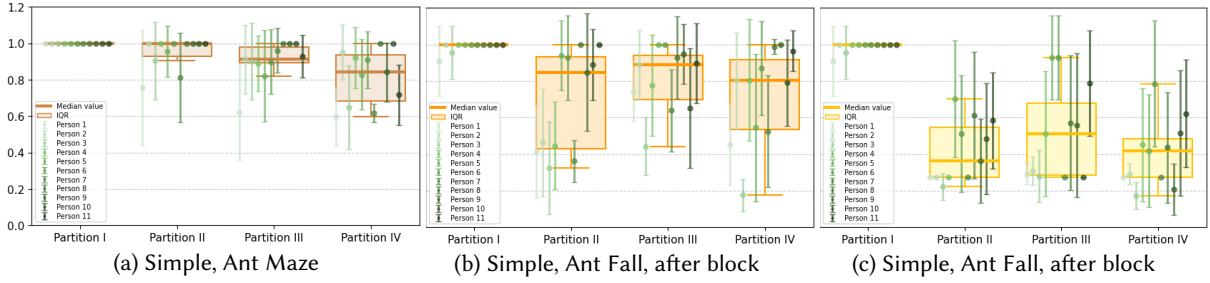
}

{\captionsetup[subfloat]{captionskip=0pt}
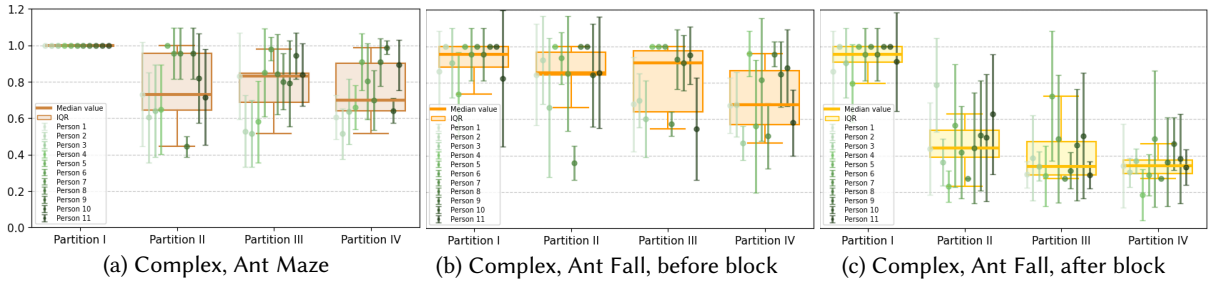
\begin{figure*}[!t]
\centering

\sbox{\boxA}{\subfloat[Complex, Ant Maze]{\includegraphics[width=2.2in]{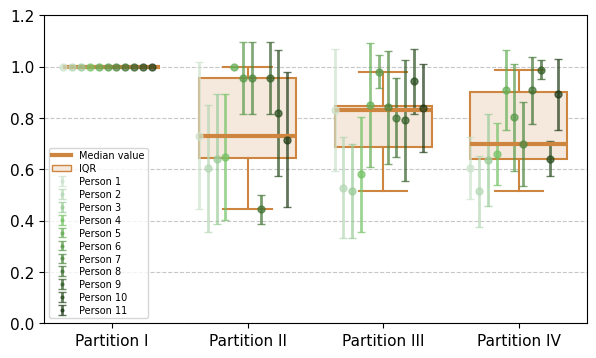}
\label{complex-AntMaze}}}
\sbox{\boxB}{\subfloat[Complex, Ant Fall, before block]{\includegraphics[width=2.2in]{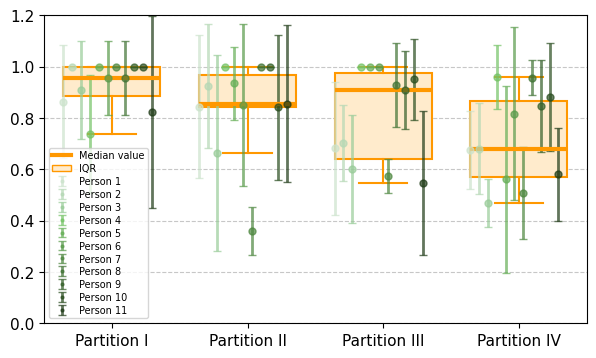}}}
\sbox{\boxC}{\subfloat[Complex, Ant Fall, after block]{\includegraphics[width=2.2in]{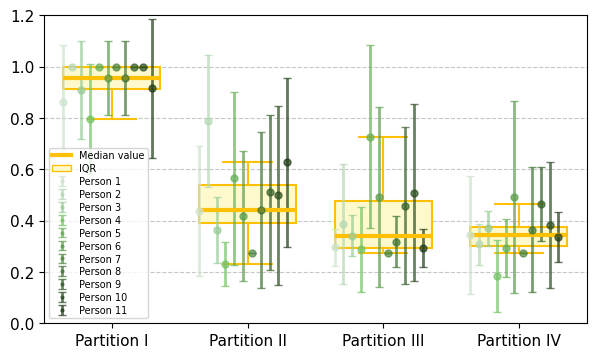}
\label{complex-AntFallAfterB}}}

\begin{tikzpicture}
  \node (C) at (10.36,0) {\usebox{\boxC}};
  \node (B) at (5.14,0) {\usebox{\boxB}};
  \node (A) at (0,-0.16) {\usebox{\boxA}};
\end{tikzpicture}
\vspace{-1.3em}
\caption{ Instructions specific to the most complex partition applying across all partitions in (a) Ant Maze task, (b) Ant Fall task, before block, (c) Ant Fall task, after block.}
\vspace{-0.7em}

\end{figure*}
}



\section{Discussion}
\label{sec:dicussion}

Our findings highlight key limitations in the translation ability and symbolic representation alignment of the LLMs in developmental learning agents. Even when instructions align with the environment's symbolic partitions, performance is often lower than expected and inconsistent across trials, suggesting that LLMs do not reliably exploit symbolic structures. Conversely, mismatched instructions sometimes yield comparable results, indicating that LLMs may rely on surface-level patterns rather than true symbolic grounding. Such behavior reflects a core issue of translation ability: although LLM-generated responses may appear syntactically correct and fluent, they do not necessarily reflect the true symbolic planning process of the agent. This disconnect undermines the use of LLMs as interpretable intermediaries in human-agent interaction, especially in safety-critical applications where understanding and verifying behavior is essential. These limitations point to a broader alignment problem between language and symbolic reasoning in current LLMs. As a result, the generated language cannot be reliably used to infer or verify the agent's decision-making process. Future works will need to address this limitation. While we believe that LLMs lack grounding, Vision-Language-Action models are bound to address these limitations in the future.

\section{Conclusion}
\label{sec:conclusion}
In this study, we examined whether large language models can effectively translate human natural language instructions into the emergent symbolic representations of developmental learning agents. Through a series of experiments on the Ant Maze and Ant Fall partitions emerged by the hierarchical reinforcement learning algorithm STAR, we demonstrated that LLMs have some ability to translate human instructions into internal symbolic representations, however it fails to understand using tools to interact with the environment. The inconsistency in G-BLEU scores and mismatched instruction-partition highlights a lack of stable alignment between human language and symbolic plans, indicating that the capacity of LLMs in alignment is inconclusive and should still be improved. 
Future research must focus on bridging this gap through enhanced grounding mechanisms to enable more reliable alignment between linguistic commands and the developmental agent's internal symbolic decision-making processes.

\bibliography{sample-ceur}

\section*{Appendix}
\appendix
\section{Prompt Design}
\label{appendix-prompt}
We take the Partition IV in Ant Maze and Ant Fall for example.
\subsection{Partition IV in Ant Maze}
\begin{multicols}{2}
    \begin{verbatim}
Data:
-State: Region 5
-Goal: Region 4

-Adjacency list:
    Region 1: [6, 7, 11, 12, 14, 18]
    Region 2: [12, 13, 14, 15]
    Region 3: [4, 19, 21]
    Region 4: [3]
    Region 5: [6, 7, 8, 9, 11]
    Region 6: [1, 5, 7, 11]
    Region 7: [1, 5, 6, 12]
    Region 8: [5, 9, 10, 12]
    Region 9: [5, 8, 10]
    Region 10: [8, 9, 12, 13]
    Region 11: [1, 5, 6, 17]
    Region 12: [1, 2, 7, 8, 10]
    Region 13: [2, 10, 15]
    Region 14: [1, 2, 15, 18]
    Region 15: [2, 13, 14, 16, 18, 20]
    Region 16: [15, 17, 18]
    Region 17: [11, 16, 18]
    Region 18: [1, 14, 15, 16, 17]
    Region 19: [3, 20, 21, 22, 23]
    Region 20: [15, 19]
    Region 21: [3, 19, 22]
    Region 22: [19, 21, 23]
    Region 23: [19, 22]

-The top-down view of the maze is shown 
 below, 'W' represents walls, 'A' 
 represents the ant's current position, 
 'G' represents the goal. The number
 represents the region number:





4  4  4  4  4  3  21 21 22 22 22 22 23 
4  4  4  4  4  3  19 19 19 19 19 19 19 
4  G  4  4  4  3  19 19 19 19 19 19 19 
W  W  W  W  W  W  W  W  W  19 19 19 19 
W  W  W  W  W  W  W  W  W  19 19 19 19 
W  W  W  W  W  W  W  W  W  19 19 19 19 
W  W  W  W  W  W  W  W  W  20 20 20 20 
10 10 10 10 10 13 13 15 15 15 15 15 15 
9  9  8  12 12 2  2  15 15 15 15 15 15 
5  5  5  7  1  14 14 15 15 15 15 15 15 
5  A  5  6  1  18 18 18 18 18 16 16 16 
5  5  5  11 11 17 17 17 17 17 17 17 17 

-Instruction:
    Move right until you completely pass the 
    wall on your left, move up until you have 
    crossed the upper wall, turn left and 
    proceed until you reach the goal.


-Thinking Process:
    1. Identify the agent's current region and
       the goal region.
    2. Interpret the Instruction: Understand 
       the directional commands provided in the
       instruction and translate them into 
       movements between regions.
    3. Plan the Route: Based on the adjacency
       list and the maze layout, determine the 
       sequence of regions the agent should 
       traverse to follow the given instructions
       and reach the goal.
    4. Check for each region of the sequence 
       if the agent can move directly to the 
       next. If not, correct the sequence
       according to the instructions. 
    \end{verbatim}
\end{multicols}
\newpage
\subsection{Partition IV in Ant Fall}
    \begin{multicols}{2}
    \begin{verbatim}
Data:
- State: Region 1
- Goal: Region 3
- Block: Region 8

- Adjacency list:
    Region 1: [16, 17, 19, 21, 22]
    Region 2: [5, 6, 11, 13]
    Region 3: [4, 24]
    Region 4: [3, 23, 24]
    Region 5: [2, 11, 12, 13]
    Region 6: [2, 11, 13, 14, 17, 19, 20, 22]
    Region 7: [8, 11, 12]
    Region 8: [7, 10, 12, 13]
    Region 9: [10, 14]
    Region 10: [8, 9, 13, 14]
    Region 11: [2, 5, 6, 7, 12]
    Region 12: [5, 7, 8, 11, 13]
    Region 13: [2, 5, 6, 8, 10, 12, 14]
    Region 14: [6, 9, 10, 13]
    Region 15: [16, 18, 21]
    Region 16: [1, 15, 21, 22]
    Region 17: [1, 6, 19]
    Region 18: [15, 20, 22]
    Region 19: [1, 6, 17, 22]
    Region 20: [6, 18, 22]
    Region 21: [1, 15, 16]
    Region 22: [1, 6, 16, 18, 19, 20]
    Region 23: [4, 24, 25]
    Region 24: [3, 4, 23, 25]
    Region 25: [23, 24]

- The top-down view of the maze is shown 
  below:

3  3  3  3  3  3  3  24 24 24 25 25 25 25 25
3  3  G  3  3  3  3  24 24 24 23 23 23 23 23
3  3  3  3  3  3  3  4  4  4  23 23 23 23 23
P  P  P  P  P  P  P  P  P  P  P  P  P  P  P
P  P  P  P  P  P  P  P  P  P  P  P  P  P  P
15 15 18 18 18 20 20 20 6  14 14 14 14 9  9
21 16 22 22 22 22 22 22 6  14 14 14 14 10 10
21 16 22 22 22 22 22 22 6  13 13 13 13 13 B
21 16 22 22 22 22 22 22 6  2  5  5  12 12 B
1  1  1  1  1  1  19 19 6  2  5  5  12 12 8
1  1  1  1  1  1  17 17 6  2  5  5  12 12 7
1  1  1  1  1  1  17 17 6  2  5  5  12 12 7
1  1  A  1  1  1  17 17 6  11 11 11 11 11 7
1  1  1  1  1  1  17 17 6  11 11 11 11 11 7

-Explanation:
    P represents pit, A represents the agent's 
    current position, B represents the movable 
    block which can be pushed by agent in four 
    directions, G represents the goal, the
    number represents the region number. The 
    block and the pit have the same width, the
    only way that the agent can pass the pit 
    is to push the block to fill the pit and 
    bridge the regions or it will not go 
    through the pit: The action push means 
    that the block moves one step in front of 
    the agent in the direction that agent moves.


-Instruction:
    Move east until reaching the end, then 
    go north to push the movable block into 
    the pit, cross the pit using the block
    as a bridge, and finally head west to
    reach the goal.

-Thinking Process:
    1. Identify the agent's current region
       and the goal region.
    2. Interpret the Instruction: Understand 
       the directional commands provided in 
       the instruction and translate them
       into movements between regions.
    3. Plan the Route: Based on the adjacency
       list, the maze layout and the 
       explanation, if the agent follows
       the instructions to reach the goal,
       describe the sequence of regions
       traversed by the agent.
    4. Check for each region of the sequence
       if the agent can move directly to the
       next. If not, correct the sequence 
       according to the instructions.
    \end{verbatim}

\end{multicols}
\newpage

\section{11 Instructions given to the partitions in AntMaze and AntFall}
\label{11instructions}

\begin{table}[htbp]
\caption{Natural language instructions for AntMaze and AntFall tasks}
\label{tab:antmaze-antfall-instructions}

\begin{tabularx}{\textwidth}{c|X|X}
\toprule
\textbf{Instructions} & \textbf{AntMaze} & \textbf{AntFall} \\
\midrule
1 & Move east until you are past the wall, then go north beyond the upper barrier, turn west, and continue until you reach the goal. & Move east until reaching the end, go north to push the block into the pit, cross the pit using the block, and finally head west to the goal. \\
\midrule
2 & Head right until there’s no obstruction in your way, then move up until the path is clear, turn left, and proceed to your destination. & Travel right as far as possible, move up to push the block forward into the pit, step onto it to cross, and then turn left to reach the goal. \\
\midrule
3 & Travel right to get around the first wall, ascend straight up to clear the second, then shift left and move toward the goal. & Head east until there’s no path ahead, move north to align with the block, push it forward into the pit, cross over it, and continue west to the goal. \\
\midrule
4 & Move horizontally to the right until you pass the boundary, then go straight up until no walls remain, turn left, and continue forward. & Move straight east until stopped, go north to push the block into the pit, walk over it, and move west until you reach the goal. \\
\midrule
5 & Proceed east to navigate around the wall, then ascend north until you are clear, turn west, and move straight to your target. & Walk east along the open path, move north to push the block into the pit, use it as a bridge to cross, and turn west to the goal. \\
\midrule
6 & Walk right until you exit the confined space, then go up beyond the vertical wall, turn left, and follow the open path to the goal. & Move right until reaching the boundary, step north to the block, push it into the pit, cross the pit safely, and proceed left to the goal. \\
\midrule
7 & Move sideways to the right until the wall is behind you, then climb upwards until there's no barrier, turn left, and walk toward the goal. & Travel east until you hit an obstacle, go up to push the block into the pit, cross over, and move left to the goal. \\
\midrule
8 & Head eastward until you escape the enclosed area, ascend northward past the last obstruction, then turn west and reach your goal. & Head east until the end, go up to push the block forward, use it to walk across the pit, and turn west to reach the goal. \\
\midrule
9 & Travel right along the open path until no wall blocks your way, go straight up past the top structure, then turn left and proceed to your destination. & Move eastward until there’s no more space, step north to push the block, let it fill the pit, cross over, and continue west to the goal. \\
\midrule
10 & Move toward the right until you have an open vertical passage, then go up until the way is clear, turn left, and walk directly to your goal. & Walk right until the path ends, move up to the block, push it forward to fill the pit, step on it to cross, and then head left to the goal. \\
\midrule
11 & Navigate eastward beyond the boundary, then ascend straight up to clear the structure, turn left, and reach the goal without further obstacles. & Travel east until the stopping point, move north to push the block, use it to bridge the pit, step over it, and finally walk west to the goal. \\
\bottomrule
\end{tabularx}

\end{table}

\end{document}